\documentclass{article}
\usepackage{amsmath,amssymb,graphicx,mlspconf}
\usepackage{booktabs}
\usepackage{xcolor}
\usepackage[ruled]{algorithm}
\usepackage[noend]{algpseudocode}
\usepackage{tikz}
\usetikzlibrary{arrows.meta,positioning}
\usepackage{needspace}

\title{Adaptive Filtering of the KV Cache: Diagnosing and Correcting Structural-Role Bias in LLM Inference}

\name{Soumil Mandal\sthanks{Accepted at the 2026 IEEE MLSP workshop, Atlanta, Georgia, USA, but not presented; proof of acceptance is included at the end of this document.}\sthanks{This work was carried out as an independent personal project during the author's time at ServiceNow, and does not represent ServiceNow's official research, products, or views.}}
\address{ServiceNow \\
         soumil.mandal@servicenow.com}

\begin{document}

\maketitle

\begin{abstract}
Attention-based KV cache eviction (H2O and its descendants) compresses the memory-constrained state of a long-context model by ranking tokens on accumulated attention mass, treated here as signal energy, and keeping the heaviest. On schema-dense input streams such as nested JSON, this score acts as a non-stationary filter that disproportionately retains noise: a non-content sink role (delimiters or whitespace) carries an order of magnitude more energy than any content role, and structural KEY tokens are over-retained at roughly $1.8\times$ the rate of the answer-carrying VALUE tokens, collapsing exact-match accuracy from 88\% to 0\% at a 5\% budget as the signal-to-noise ratio of the retained state degrades. A counterfactual experiment establishes that suppressing KEY tokens is the best deployable filter. Our retraining-free, role-conditional allocation over SnapKV's windowed score, governed by a single \emph{tuned} hyperparameter, closes 63--98\% of the H2O gap at sub-20\% budgets and, at higher budgets, modestly matches or exceeds full-cache accuracy, a small, seed-sensitive denoising effect (borderline significant at $B{=}0.50$; not distinguishable from zero at $B{=}0.30$ over four seeds). A 15\,MB linear role probe supplies these labels at negligible inference cost, though matching parser-level downstream accuracy remains open.
\end{abstract}
\section{Introduction}
\label{sec:intro}

KV cache eviction is the dominant lever for memory and latency in long-context LLM inference. The canonical method, H2O~\cite{zhang2023h2o}, ranks tokens by accumulated attention mass and keeps a fixed fraction (the ``budget'') of heavy hitters, training-free, drop-in, grounded in the observation that a few tokens receive disproportionate attention. Through a signal-processing lens, the cache is a memory-constrained representation of a streaming signal and eviction is an adaptive filter retaining the informative components under a hard rate budget.

We study a regime where this scoring assumption fails. \emph{Schema-dense long contexts} (deeply nested JSON, XML, large markdown tables) are increasingly common in retrieval-augmented and structured-data agent workloads yet absent from standard KV-compression benchmarks. Here an order-of-magnitude concentration of attention mass on \emph{structural} tokens (delimiters, whitespace, JSON keys) decouples ``attention mass'' (signal energy) from ``answer relevance'' (information content): H2O's heavy-hitter filter optimizes the wrong criterion, retaining high-energy structural noise, and exact-match (EM) accuracy on leaf-value QA collapses from $88\%$ at full budget to $0\%$ at a $5\%$ budget.

This paper makes four contributions. (i) We \emph{diagnose} a structural-routing bias in H2O on schema-dense contexts: an attention-sink role accumulates roughly $30\times$ the mean per-token mass of any content role, and KEY tokens are over-retained at $1.8$--$1.9\times$ the rate of VALUE tokens. (ii) A \emph{counterfactual} experiment that bans individual roles from retention shows that evicting every KEY token is the best deployable policy, while delimiters and values are both required scaffold. (iii) A composable \emph{role-conditional allocator} on top of SnapKV~\cite{li2024snapkv} composes super-additively; at higher budgets it matches or modestly exceeds full-cache accuracy (a small, seed-sensitive effect). (iv) We replicate across four model families (the claims split into a universal core and a model-coupled residual) and study a 15\,MB linear role probe as a parser substitute.

\section{Background and related work}
\label{sec:related}

\textbf{The H2O family.} H2O~\cite{zhang2023h2o} accumulates a heavy-hitter score $s(t)=\sum_q p_t(q)$ during prefill and keeps the top-$k$ tokens per (layer, head). Descendants include Scissorhands~\cite{liu2023scissorhands}, FastGen~\cite{ge2024fastgen}, SnapKV~\cite{li2024snapkv} (windowed mass over the last $W$ queries with a 1-D max-pool over the KV axis), and PyramidKV~\cite{cai2024pyramidkv}. Later methods refine \emph{where} budget is spent: an attention-variance term (CAKE~\cite{qin2025cake}), query-aware selection (Quest~\cite{tang2024quest}), coarse-to-fine cascades (RocketKV~\cite{behnam2025rocketkv}), one-shot self-attention top-$k$ (SAGE-KV~\cite{wang2025sagekv}), a second-order criterion (OBCache~\cite{gu2025obcache}), or make eviction \emph{learned} (KVP~\cite{moschella2026kvp}, SideQuest~\cite{kariyappa2026sidequest}). Our allocator stays training-free and orthogonal: it reallocates an existing score across structural roles rather than proposing a new one, extending the sink from position $0$ to hundreds of distributed delimiter positions. StreamingLLM~\cite{xiao2024streamingllm} adds a fixed BOS-sink slot, which~\cite{sun2024massive,gu2025attentionsink,barbero2025firsttoken} tie to massive activations and~\cite{ruscio2025sinking} casts as a geometric reference-frame effect.

\textbf{Budget allocation.} Ada-KV~\cite{feng2025adakv} and HeadKV~\cite{fu2025headkv} allocate a global budget non-uniformly across attention \emph{heads}; we partition along a structural-role axis within a head, and the two could be stacked. The head view is grounded in retrieval heads~\cite{wu2024retrieval}, which CompressKV~\cite{lin2025compresskv} uses to flag unimportant tokens.

\textbf{Closest prior art.} \cite{chen2025pitfalls} diagnose an eviction bias at the granularity of instruction blocks and propose equal-rate ``fair eviction.'' We work at the finer granularity of structural roles and find the \emph{opposite} policy direction: structural QA benefits from retention biased \emph{away} from KEY tokens. Benchmarks such as LongBench~\cite{bai2024longbench} and RULER~\cite{hsieh2024ruler} dominate the literature but do not stress deeply schema-dense leaf-value QA.

\section{Diagnosis: a structural-routing bias}
\label{sec:diagnosis}

\textbf{Setup.} We evaluate on four synthetic corpora: indented \texttt{synthetic\_json} (nested objects, leaf-value queries), \texttt{synthetic\_json\_compact} (no whitespace), \texttt{synthetic\_xml} (same content as XML), and \texttt{wikitable} (markdown tables with distractors). Accuracy is exact-match (EM) over $n{=}50$ prompts at seed $0$ across budgets $B\in\{5,10,20,30,50,75,100\}\%$ ($B{=}100\%$ is the no-eviction full-cache baseline). Prefill records per-(layer, head) attention mass; eviction installs a keep-mask before an explicit-scores decode path. Runs use fp16 on one 80\,GB A100; ctx-16k numbers reproduce on fp32, ruling out a decode-path artifact.

We first profile attention on \texttt{Llama-\allowbreak3.1-\allowbreak8B-\allowbreak Instruct} (fp16) across 604 prompts in three context buckets (8k/\allowbreak16k/\allowbreak32k), labelling each token with a structural role in $\{$KEY, VALUE, DELIM, HEADER, PROSE, WS$\}$ (KEY/\allowbreak HEADER schema labels, DELIM punctuation, WS whitespace).

\begin{figure}[t]
\centering
\includegraphics[width=0.98\linewidth]{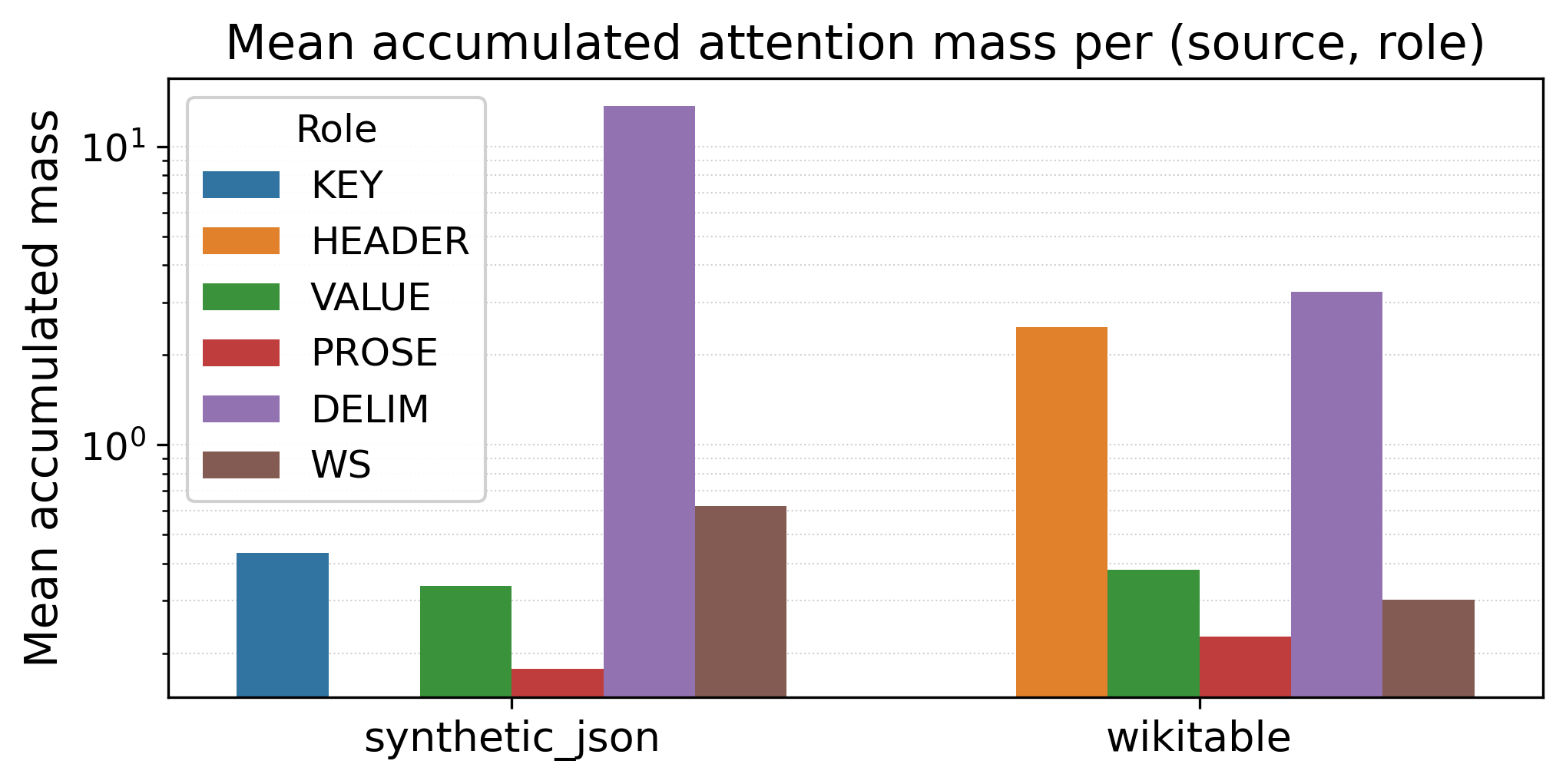}
\caption{Per-role mean accumulated attention mass on \texttt{synthetic\_json} (Llama-3.1-8B, log axis). The DELIM sink exceeds every content role by an order of magnitude; a linear axis would hide all non-DELIM roles. This is the structural sink that H2O's heavy-hitter score conflates with answer relevance.}
\label{fig:diag}
\end{figure}

\looseness=-1
\textbf{Mass concentration.} On \texttt{synthetic\_json}, mean per-token mass on DELIM is $13.64$, against $0.434$ (KEY), $0.336$ (VALUE), $0.177$ (PROSE), $0.623$ (WS): an order-of-magnitude structural sink (Fig.~\ref{fig:diag}). As signal energy, the sink carries $\sim$$40\times$ the VALUE signal's per-token energy ($13.64$ vs.\ $0.336$), so an energy-based filter locks onto the noise floor.

\textbf{Retention bias.} Table~\ref{tab:q2} reports each role's H2O retention rate. The KEY-vs-VALUE ratio is $1.94/1.79/1.65\times$ at $B{=}5/10/20\%$ on JSON, and HEADER retention on \texttt{wikitable} reaches $47\%$ at $B{=}5\%$, while answer-carrying VALUE is systematically under-retained. Mass-quartile bucketing confirms the mechanism: VALUE concentrates in the bottom (lowest-mass) quartile and KEY in the top, so the scorer parks answer tokens where they are evicted first. The bias is layer-pervasive but not uniform: layer 0 inverts it (VALUE $>$ KEY), and on \texttt{wikitable} layer 1 reaches a $34\times$ HEADER-over-VALUE mass ratio.

\textbf{The sink is not the BOS sink.} Splitting the sink into the BOS position and non-BOS DELIM shows that BOS is exactly StreamingLLM's fixed slot (one position, retained almost always), whereas the operational sink is $\sim$674 distinct DELIM positions per prompt at text-inferable locations. A single fixed slot cannot capture a distributed, format-dependent sink; role-conditional allocation can.

\textbf{A retained-state SNR estimate.} The filtering language admits a coarse number. Treating attention mass as per-token energy, a retained-state SNR is the kept energy on VALUE tokens relative to structural DELIM/KEY. On the raw \texttt{synthetic\_json} state the VALUE-to-DELIM ratio is $0.336/13.64\approx0.025$ ($\approx{-}16$\,dB): the answer already sits an order of magnitude below the structural noise floor. Since H2O retains DELIM and KEY at higher rates than VALUE (Table~\ref{tab:q2}), eviction pushes this ratio down as the budget tightens, and EM tracks it ($88\%\!\to\!0\%$ as $B{:}\,100\%\!\to\!5\%$). Role-conditional allocation (\S\ref{sec:combined}) is thus an SNR-raising filter: capping noise roles and floor-protecting VALUE lifts the ratio, recovering EM. We report it as a directional estimator, not a formal control.

\begin{table}[t]
\centering
\caption{H2O retention rate per (corpus, role, budget) on Llama-3.1-8B. KEY/HEADER are over-retained and VALUE under-retained relative to the random baseline $B$.}
\label{tab:q2}
\small
\begin{tabular}{llrrr}
\toprule
Corpus & Role & $B{=}5\%$ & $B{=}10\%$ & $B{=}20\%$ \\
\midrule
\texttt{json} & KEY    & \textbf{0.060} & \textbf{0.121} & \textbf{0.242} \\
\texttt{json} & VALUE  & 0.032 & 0.068 & 0.147 \\
\texttt{json} & DELIM  & 0.105 & 0.185 & 0.315 \\
\texttt{json} & PROSE  & 0.027 & 0.049 & 0.096 \\
\texttt{wikitable} & HEADER & \textbf{0.467} & \textbf{0.616} & \textbf{0.768} \\
\texttt{wikitable} & VALUE  & 0.051 & 0.101 & 0.200 \\
\bottomrule
\end{tabular}
\end{table}

\section{Counterfactual eviction and the combined method}
\label{sec:method}

\textbf{Counterfactual.} Correlation between mass and role does not establish causation. For each (layer, head, prompt) we select the H2O top-$k$ from a candidate set excluding all tokens of one role, then decode (Table~\ref{tab:cf}). On \texttt{synthetic\_json} ctx 8k ($n{=}50$), H2O collapses to $0\%$ below $B{=}0.5$, whereas \texttt{No-KEY} recovers to $0.82$ at $B{=}0.5$ and hits the $0.88$ ceiling at $B{=}0.75$, using strictly less candidate budget. \texttt{No-DELIM} and \texttt{No-VALUE} collapse to $0\%$ at every sub-$1.0$ budget, so delimiters and values are both required scaffold even though delimiters carry no schema-level answer. The retention ordering is therefore DELIM $\approx$ VALUE $\gg$ KEY, inverting the naive ``keys index, values are content'' reading.

\begin{table}[t]
\centering
\caption{Counterfactual role-eviction EM on indented JSON (ctx 8k, $n{=}50$). Each \texttt{No-X} condition evicts every token of role X and fills the rest by mass. \texttt{No-KEY} reaches the full-cache baseline ($0.88$) using less budget; \texttt{No-DELIM} is uniformly catastrophic.}
\label{tab:cf}
\small
\begin{tabular}{lrrrr}
\toprule
$B$ & All (H2O) & No-DELIM & \textbf{No-KEY} & No-VALUE \\
\midrule
0.20 & 0.00 & 0.00 & \textbf{0.14} & 0.00 \\
0.30 & 0.00 & 0.00 & \textbf{0.32} & 0.00 \\
0.50 & 0.14 & 0.00 & \textbf{0.82} & 0.00 \\
0.75 & 0.48 & 0.00 & \textbf{0.88} & 0.28 \\
1.00 & 0.88 & 0.88 & 0.88 & 0.88 \\
\bottomrule
\end{tabular}
\end{table}

\textbf{SnapKV mechanism.} SnapKV replaces $\sum_q p_t(q)$ with a windowed sum over the last $W$ queries, then a 1-D max-pool (kernel $7$) over the KV axis. At $W{=}64$, windowed mass alone closes $81\%$ of the H2O gap at $B{=}0.5$; adding the pool closes $100\%$ at ctx 16k. The pool down-weights \emph{isolated} high-mass tokens (single sinks) and up-weights content \emph{clusters}, so SnapKV's gain is largely implicit sink suppression, not recency. Sweeping $k\in\{3,5,7,11,15\}$, EM rises monotonically in $k$ at binding budgets ($0.06\!\to\!0.54$ at $B{=}0.05$), so the default $k{=}7$ is conservative here; disabling the pool collapses low-budget EM to $0.16$.

\subsection{The combined method}
\label{sec:combined}

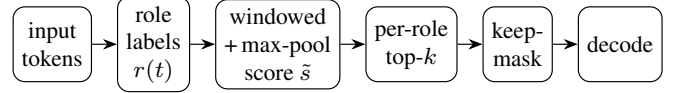
\begin{figure}[t]
\centering
\resizebox{\columnwidth}{!}{%
\begin{tikzpicture}[
  box/.style={draw, rounded corners, align=center, inner sep=3pt, font=\footnotesize, minimum height=9mm},
  >=Stealth, node distance=3mm]
\node[box] (a) {input\\tokens};
\node[box, right=of a] (b) {role\\labels\\$r(t)$};
\node[box, right=of b] (c) {windowed\\{+}\,max-pool\\score $\tilde s$};
\node[box, right=of c] (d) {per-role\\top-$k$};
\node[box, right=of d] (e) {keep-\\mask};
\node[box, right=of e] (f) {decode};
\draw[->] (a)--(b); \draw[->] (b)--(c); \draw[->] (c)--(d);
\draw[->] (d)--(e); \draw[->] (e)--(f);
\end{tikzpicture}}
\caption{Inference pipeline of the combined method. Role labels $r(t)$ come from a structural parser in our accuracy experiments; \S\ref{sec:deploy} studies replacing it with a 15\,MB linear probe (currently an imperfect substitute). Scoring and pooling are SnapKV; the per-role top-$k$ and keep-mask are the role-conditional allocator.}
\label{fig:pipeline}
\end{figure}

The two findings address orthogonal failure modes, position bias (the sink) and content bias (KEY over-retention), so we compose them into a single role-conditional allocator (Algorithm~\ref{alg:combined}, Fig.~\ref{fig:pipeline}): score by windowed mass, smooth with the SnapKV max-pool, then keep the top-$\lceil \alpha_r B\, n_{\text{kv}}\rceil$ tokens \emph{within each role bucket} $r$ of each (layer, head), redistributing slack from saturated buckets, and decode through the keep-mask. The role policy $\alpha=\{V{:}0.70, D{:}0.20, P{:}0.05, W{:}0.05, \text{KEY}{:}\alpha_{\text{KEY}}\}$ is renormalized per bucket.

\begin{algorithm}[t]
\caption{Role-conditional KV allocation (per layer, per head).}
\label{alg:combined}
\small
\begin{algorithmic}[1]
\Require roles $r(t)$; budget $B$; window $W{=}64$; kernel $k{=}7$; weights $\alpha$
\State $s(t)\gets\sum_{q\in\text{last }W}p_t(q)$ \Comment{windowed mass (SnapKV)}
\State $\tilde s(t)\gets \mathrm{MaxPool}_k\!\big(s(t)\big)$ \Comment{1-D pool, KV axis}
\ForAll{role buckets $r$}
  \State keep top-$\lceil \alpha_r B\,n_{\text{kv}}\rceil$ tokens of $r$ by $\tilde s$
\EndFor
\State redistribute unused budget from saturated buckets
\State \Return keep-mask; decode through it
\end{algorithmic}
\end{algorithm}

\textbf{Hyperparameters.} The method exposes \emph{one tuned knob over three diagnosis-fixed constants}, not a bank of free parameters. Fixed, not tuned: the window $W{=}64$ and pool kernel $k{=}7$ are SnapKV defaults (the kernel sweep above shows EM monotone in $k$, so $k{=}7$ is conservative), and the content-role weights $\{V{:}0.70, D{:}0.20, P{:}0.05, W{:}0.05\}$ are read off the counterfactual ordering DELIM$\approx$VALUE$\gg$KEY (Table~\ref{tab:cf}) rather than swept. Tuned: only $\alpha_{\text{KEY}}\in[0,0.2]$, whose per-cell optimum is stable across all five budgets. A $\pm0.1$ sweep of the fixed content-role weights (ctx 16k, $B\in\{0.05,0.30,0.50\}$, $n{=}50$) confirms the split is robust: EM moves by at most $4$ EM points at any budget, \emph{except} when a weight is driven to exactly zero. Zeroing WS costs $-12$ EM at $B{=}0.30$; zeroing PROSE collapses EM to $0.00$ at every budget, because the parser labels the question text as PROSE and evicting it removes the model's only retrieval cue. KEY is the one role the method is designed to zero; the other three carry a strictly-positive, load-bearing floor, which the fixed point respects.

\textbf{Results.} Table~\ref{tab:combined} compares the combined method ($\alpha_{\text{KEY}}{=}0$) against SnapKV-only, the allocator alone, and H2O; it dominates each at every budget. At ctx 16k, $B{=}0.05$, SnapKV alone closes $21\%$ of the H2O gap and the allocator alone $0\%$, yet the combination closes $63\%$ (a factor-of-three super-additivity), rising to $86\%$ and $98\%$ at $B{=}10/20\%$. The gain is context-robust but bounded: the combined-vs-SnapKV margin at $B{=}5\%$ is $+0.16/{+}0.36/{+}0.18$ at ctx 8k/16k/32k, peaking near 16k.

\needspace{8\baselineskip}
\vspace{-2pt}
\textbf{Fair-eviction baseline.} We implement the fair-eviction rule of~\cite{chen2025pitfalls} with our structural-role partition in place of their instruction-block one (equal retention rate per role). On JSON it is \emph{worse than vanilla SnapKV} ($-0.18$ to $-0.70$ EM) and loses to the best role-conditional policy in every cell. The same partition-and-allocate container thus yields opposite optima on opposite workloads: equal rate for instruction-following, retention biased away from KEY for structural QA.
\vspace{-2pt}

\textbf{Query-aware baseline (Quest).} Quest~\cite{tang2024quest} re-selects KV pages at \emph{every} decode step; our harness makes one static post-prefill decision, so we implement a one-shot analogue matched to that regime (page-level key-bound scoring against the mean query of the same $W{=}64$ window). It trails both H2O/SnapKV and the combined method at every budget (EM $0.06/0.28/0.48/0.68$ at $B{=}5/20/30/50\%$ vs.\ $0.56/0.86/0.98/0.96$). Omitting Quest's per-step re-selection, this is a \emph{lower bound}, not a head-to-head comparison; a full incremental reimplementation is future work. It is consistent with our framing that query-aware selection alone does not resolve the role/content bias.

\begin{table}[t]
\centering
\caption{Combined SnapKV $\times$ role-conditional (\texttt{no-key}, $\alpha_{\text{KEY}}{=}0$) vs.\ its ingredients on \texttt{synthetic\_json} ctx 16k ($n{=}50$, seed 0, full-cache baseline $0.88$). The $B{\geq}0.30$ margin over the full-cache baseline is seed-sensitive (Table~\ref{tab:seeds}).}
\label{tab:combined}
\small
\begin{tabular}{lrrrr}
\toprule
$B$ & H2O & no-key & SnapKV & \textbf{Combined} \\
\midrule
0.05 & 0.02 & 0.02 & 0.20 & \textbf{0.56} \\
0.10 & 0.02 & 0.04 & 0.34 & \textbf{0.76} \\
0.20 & 0.00 & 0.18 & 0.70 & \textbf{0.86} \\
0.30 & 0.02 & 0.34 & 0.80 & \textbf{0.98} \\
0.50 & 0.22 & 0.86 & 0.88 & \textbf{0.96} \\
\bottomrule
\end{tabular}
\end{table}

\textbf{KV eviction as a quality lever (seed-sensitive).} At ctx 16k, $B\in\{0.30,0.50\}$, the combined method reaches $0.98$ and $0.96$ on the seed-0 draw used elsewhere, above the full-cache baseline ($0.88$) by $10$ and $8$ EM. This margin is fragile: over three more prompt draws (Table~\ref{tab:seeds}, $n{=}200$ pooled/budget) it shrinks sharply. At $B{=}0.30$ the mean is $+0.01$ with a $95\%$ CI of $[{-}0.03,{+}0.07]$ straddling zero (McNemar $p{\approx}0.80$; $7$ regressions vs.\ $9$ recoveries), so the seed-0 ``zero regressions'' pattern is a lucky draw, not a property. At $B{=}0.50$ a small effect survives ($+3.5$ EM, $p{\approx}0.046$; $1$ regression vs.\ $8$). The mechanism is plausible (suppressing high-attention KEY tokens denoises the focus on the leaf VALUE), but the defensible claim is only that at higher budgets the method \emph{matches or modestly exceeds} the full cache, not that eviction is a reliable quality lever below $B{=}0.50$.

\begin{table}[t]
\centering
\caption{Multi-seed robustness of the over-baseline margin ($\Delta$EM $=$ combined method $-$ full-cache baseline), \texttt{synthetic\_json} ctx 16k, $n{=}50$ prompts/seed. Only seed 0 (the draw used elsewhere in this paper) shows a large margin; pooled over four seeds the $B{=}0.30$ effect is indistinguishable from zero and the $B{=}0.50$ effect is small and borderline.}
\label{tab:seeds}
\small
\begin{tabular}{lrr}
\toprule
Seed / statistic & $\Delta$EM @ $B{=}0.30$ & $\Delta$EM @ $B{=}0.50$ \\
\midrule
seed 0           & $+0.10$ & $+0.08$ \\
seed 1           & $-0.02$ & $+0.02$ \\
seed 2           & $\phantom{+}0.00$ & $+0.04$ \\
seed 3           & $-0.04$ & $\phantom{+}0.00$ \\
\midrule
mean             & $+0.01$ & $+0.035$ \\
95\% CI          & $[-0.03,{+}0.07]$ & $[{+}0.01,{+}0.065]$ \\
McNemar $p$      & $0.80$ & $0.046$ \\
\bottomrule
\end{tabular}
\end{table}

\textbf{The $\alpha$ family.} Sweeping $\alpha_{\text{KEY}}\in\{0,0.02,0.05,0.1,0.2\}$, the optimum is bimodal: $\alpha_{\text{KEY}}{=}0$ wins on cleanly separable indented JSON at long context ($+0.08$ to $+0.42$ EM over SnapKV at ctx 16k), while $\alpha_{\text{KEY}}{=}0.02$ is the safe default elsewhere ($+0.04$ to $+0.44$ on compact JSON and XML; near-tied on \texttt{wikitable}). On compact JSON and XML, Llama's BPE merges delimiter and key characters into single tokens, so $\alpha_{\text{KEY}}{=}0$ evicts merged delimiters and fails ($0/50$ EM at every budget); the soft floor $\alpha_{\text{KEY}}{=}0.02$ preserves them. A precedence-flip ablation (relabelling merged tokens as DELIM) rescues compact JSON but not XML, so only the former collapse is a labeling artifact. Within each cell the best $\alpha$ is stable across all five budgets.

\section{Cross-model generalization}
\label{sec:cross}

We replicate the diagnosis, counterfactual, and combined method on \texttt{Mistral-\allowbreak7B-\allowbreak Instruct-\allowbreak v0.3}, \texttt{Phi-\allowbreak3-mini} (both SentencePiece), and \texttt{Qwen2.5-\allowbreak7B-\allowbreak Instruct} (a divergent tiktoken-family BPE), all on \texttt{synthetic\_\allowbreak json} ctx 16k with identical prompts ($n{=}50$). The claims split into a \emph{universal core} and a \emph{model-coupled residual} (Table~\ref{tab:grid}).

The KEY-over-VALUE retention bias (C1), the counterfactual KEY recovery (C2), and the max-pool mechanism (C7) hold on all four families: the C1 ratio at $B{=}5\%$ is $1.94/1.82/1.98/1.75\times$, and \texttt{No-KEY} lifts EM above each model's own full-cache baseline (Mistral $0.64$ vs.\ $0.10$ for \texttt{All} at $B{=}0.5$; Phi-3 $0.62$ above its $0.60$ full-cache baseline; Qwen $+0.38$ EM at $B{=}0.5$). The super-additive composition (C8) and the over-baseline effect (C9), however, are full on Llama, compressed on Mistral, and absent on Phi-3 and Qwen; C9 in particular is seed-sensitive even on Llama (Table~\ref{tab:seeds}), so we treat it as a qualified secondary observation, not a headline claim.

The sink's \emph{identity} is tokenizer-coupled but its eviction behavior is not: the dominant per-token sink lands on a different role per family (DELIM on Llama, whitespace on Mistral, none on Phi-3, PROSE on Qwen), yet the H2O ordering DELIM $\gg$ KEY $\approx$ WS $>$ VALUE $>$ PROSE is identical on all four. Phi-3 is the decisive control: it shares Mistral's tokenizer with identical char-span counts yet patterns with Qwen, falsifying a ``degrades with tokenizer divergence'' reading. The surviving mechanism is that SnapKV's max-pool already cancels the KEY bias \emph{completely} on Phi-3 and Qwen, leaving no residual for the \texttt{no-key} correction, whereas on Llama (and, smaller, Mistral) a residual remains; the low-budget combined-vs-SnapKV gain at $B{=}5\%$ makes it concrete: $+0.36$ (Llama), $+0.24$ (Mistral), $+0.00$ (Qwen).

The bias predates instruction tuning: the non-instruct \texttt{Llama-3.1-8B} base reproduces the diagnosis at near-identical magnitude (DELIM sink mass $13.54$; C1 ratio $1.91\times$ at $B{=}5\%$), so it is a property of the pretrained transformer, not an alignment artifact. Accuracy claims stay scoped to instruct models, as the base floors at $0.00$ EM even at full budget (a capability floor, not a refutation).

The $\alpha_{\text{KEY}}{=}0$ binary collapse on merged-token corpora is itself Llama-specific ($\alpha{=}0$ stays functional at $0.32$--$0.50$ EM on Mistral and Qwen), confirming the soft floor $\alpha_{\text{KEY}}{=}0.02$ as the safe cross-model default.

\textbf{Role-density boundary.} The policy gain is bounded by role density. On \texttt{wikitable} the over-attended role is HEADER at only $1.6\%$ of tokens (vs.\ JSON KEY at $\sim$$42\%$), and all five $\alpha$ curves bunch in a $0.55$--$0.66$ EM band indistinguishable from SnapKV: excluding $1.6\%$ of tokens frees too little budget to matter. The \emph{diagnosis} (the No-DELIM catastrophe) still replicates, so it is the policy gain, not the phenomenon, that is density-bounded.

\begin{table}[htbp]
\centering
\caption{Claim $\times$ model-family partition. $\checkmark$ replicates; $\times$ does not; $\circ$ partial. The top three rows are the universal core; C8/C9 are the model-coupled residual.}
\label{tab:grid}
\small
\setlength{\tabcolsep}{4pt}
\resizebox{\columnwidth}{!}{%
\begin{tabular}{lcccc}
\toprule
Claim & Llama & Mistral & Phi-3 & Qwen \\
\midrule
C1: over-retains KEY vs.\ VALUE & $\checkmark$ & $\checkmark$ & $\checkmark$ & $\checkmark$ \\
C2: ban-KEY recovers accuracy   & $\checkmark$ & $\checkmark$ & $\checkmark$ & $\checkmark$ \\
C7: max-pool load-bearing       & $\checkmark$ & $\checkmark$ & $\checkmark$ & $\checkmark$ \\
C8: super-additive comp.        & $\checkmark$ & $\checkmark$ & $\times$ & $\times$ \\
C9: over-baseline lever           & $\checkmark$ & $\checkmark$ & $\times$ & $\times$ \\
\bottomrule
\end{tabular}}
\end{table}

\section{Deployment: a linear role probe}
\label{sec:deploy}

The method needs per-token role labels, supplied by a structural parser in every experiment above. Toward parser-free deployment we train a multinomial logistic regression over a 5-token window of token IDs (current $\pm 2$): $\sim$3.85M parameters ($\sim$15\,MB at fp32), trained in $8.6$\,s on 200 prompts. \emph{In isolation} it labels roles near-perfectly (held-out accuracy $0.9952$, KEY F1 $0.9991$).

\textbf{But the probe is not a drop-in replacement.} Swapping probe for parser labels in the combined method drops downstream EM by $18$--$32$ points at ctx 16k ($0.98\!\to\!0.66$ at $B{=}0.30$). The cause is identifiable: as a purely local classifier the probe has no notion of which \emph{block} a token sits in, so it labels the question's path reference (e.g.\ \texttt{...\_83.delta\_26...}) as KEY on token shape, whereas the parser's block-level dispatch scopes that free-form text as PROSE. Under $\alpha_{\text{KEY}}{=}0$ these mislabeled question tokens get zero budget and are evicted, deleting the model's only retrieval cue (the same failure as zeroing PROSE, \S\ref{sec:combined}). Near-perfect isolated F1 thus does not imply deployment readiness; adding a block/segment feature is the natural fix, left to future work.

\textbf{Systems cost.} Probe inference is negligible: $0.04$\,s/prompt vs.\ $1.94$\,s for the parser, under $1\%$ of a $32$-token decode. The SnapKV/H2O windowed-attention recorder is a real one-time cost ($\sim$$12.9$\,s/prompt at ctx 16k). We claim \emph{no} measured memory or latency savings: to isolate the accuracy effect of an exact keep-pattern, our harness masks attention to evicted keys rather than compacting the KV tensors, so peak memory is flat across $B\in\{0.05,\dots,1.0\}$ (${<}0.1\%$) and decode is no faster. The $1{-}B$ reduction holds mechanically for any top-$k$ scheme in a compacting implementation, but we do not demonstrate it here.

\section{Conclusion}
\label{sec:conc}

Attention-mass scoring conflates signal energy (high on structural sinks) with information content (high on answer-relevant tokens), so heavy-hitter eviction is a filter matched to the wrong criterion. The contribution is three composable pieces: a diagnostic for \emph{why} recency-based methods help (implicit sink avoidance), an isolation of \emph{which} part of SnapKV does the work (the max-pool over the KV axis, not the recency floor), and a role-conditional allocator that fits on any score under a single tuned hyperparameter (three role weights fixed by the diagnosis). Together they close most of the H2O gap on schema-dense QA and compose super-additively; at higher budgets the method matches or modestly exceeds full-cache accuracy, a small, seed-sensitive effect (\S\ref{sec:combined}). It is retraining-free and, partitioning budget along the role axis \emph{within} a head, complementary to head-axis allocators (Ada-KV~\cite{feng2025adakv}, HeadKV~\cite{fu2025headkv}). Limitations: a role-density floor on sparse-role corpora (e.g.\ wikitable); the seed-sensitive over-baseline margin; a masking-based harness that isolates accuracy but does not realize the $1{-}B$ memory/latency benefit; a Quest comparison that omits per-step re-selection (a lower bound); and fp16-only runs. Adding block-context features to the probe (a naive local-window probe transfers imperfectly downstream), extending to 4-bit KV quantization, and validating on real schema-dense corpora (RAG, API logs, LongBench-style table QA) are future work. Framed in the signal-processing terms of \S\ref{sec:intro}, the allocator leaves the underlying attention-mass score untouched and reshapes only the read-out policy applied to it.

\let\oldthebibliography\thebibliography
\renewcommand{\thebibliography}[1]{%
  \oldthebibliography{#1}%
  \setlength{\itemsep}{0pt}%
  \setlength{\parsep}{0pt}}
\bibliographystyle{IEEEbib}
\bibliography{refs}

\clearpage
\onecolumn
\begin{center}
\textit{proof of acceptance}\\[1em]
\includegraphics[width=0.85\textwidth]{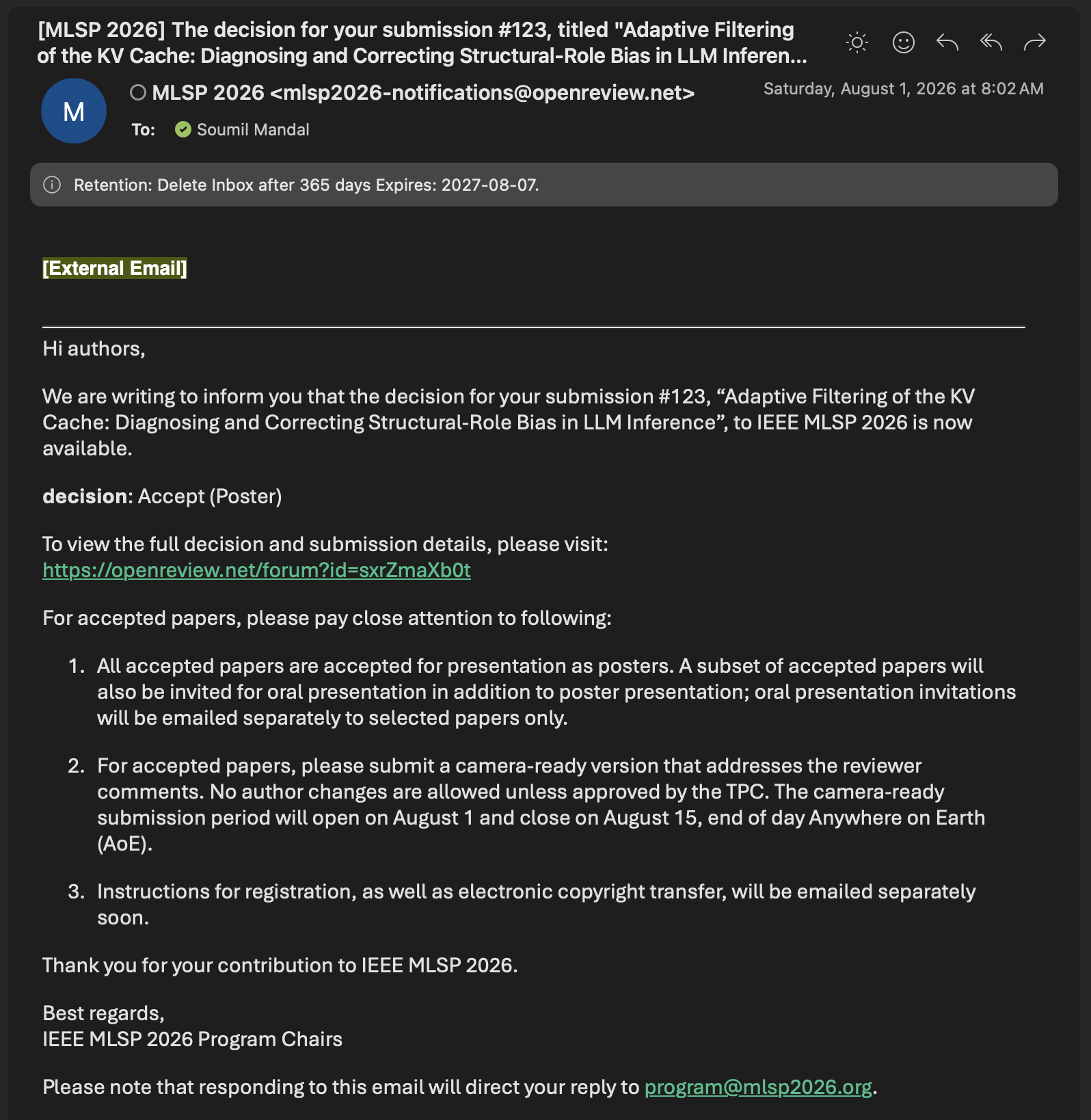}
\end{center}

\end{document}